\documentclass{article}

\pdfoutput=1

\usepackage{amsmath}
\usepackage{amsfonts}
\usepackage{mathtools}

\usepackage{natbib}

\usepackage{float}
\usepackage{booktabs}

\usepackage{hyperref}

\usepackage{graphicx} 
\graphicspath{ {figures/} }

\usepackage[margin=1in]{geometry}

\title{A comparative analysis of machine learning models in SHAP analysis}
\author{Justin Lin\footnote{Corresponding author, linjus@iu.edu, Department of Mathematics, Indiana University} and Julia Fukuyama\footnote{Department of Statistics, Indiana University} \\ for the Alzheimer’s Disease Neuroimaging Initiative\footnote{Data used in preparation of this article were obtained from the Alzheimer’s Disease Neuroimaging Initiative (ADNI) database (adni.loni.usc.edu). As such, the investigators within the ADNI contributed to the design and implementation of ADNI and/or provided data but did not participate in analysis or writing of this report. A complete listing of ADNI investigators can be found at: http://adni.loni.usc.edu/wp-content/uploads/how\_to\_apply/ADNI\_Acknowledgement\_List.pdf}}
\date{}

\begin{document}
\maketitle


\begin{abstract}
In this growing age of data and technology, large black-box models are becoming the norm due to their ability to handle vast amounts of data and learn incredibly complex data patterns. The deficiency of these methods, however, is their inability to explain the prediction process, making them untrustworthy and their use precarious in high-stakes situations. SHapley Additive exPlanations (SHAP) analysis is an explainable AI method growing in popularity for its ability to explain model predictions in terms of the original features. For each sample and feature in the data set, an associated SHAP value quantifies the contribution of that feature to the prediction of that sample. Analysis of these SHAP values provides valuable insight into the model's decision-making process, which can be leveraged to create data-driven solutions. The interpretation of these SHAP values, however, is model-dependent, so there does not exist a universal analysis procedure. To aid in these efforts, we present a detailed investigation of SHAP analysis across various machine learning models and data sets. In uncovering the details and nuance behind SHAP analysis, we hope to empower analysts in this less-explored territory. We also present a novel generalization of the waterfall plot to the multi-classification problem.
\end{abstract}

\clearpage


\section{Introduction}

Supervised learning is a machine learning paradigm in which the goal is to predict outputs given known input-output pairs. Examples include regression, in which the output is one-dimensional and continuous, or classification, in which the output is nominal. Given the advances in data collection and computing power, increasingly complex methods are being developed to handle supervised data, many of which are known as black-box methods. As the name suggests, the inner workings of these models are nebulous due to their high levels of complexity. However, they are among the highest performers when it comes to supervised learning.

In efforts to dissect and better understand these methods, explainable AI (XAI) methods are being developed to extract meaning from these so-called black-box models. One of which, SHapley Additive exPlanations (SHAP, \citet{shap}) analysis, is growing in popularity due to its ability to explain model predictions in terms of the original features. By linking predictions back to the original features, we can better understand why each sample receives the prediction it does and draw inference from these explanations.

For each sample and feature in the data, we calculate a corresponding SHAP value that quantifies the contribution of that feature to the prediction of that sample. In this way, the SHAP values encode the model's decision-making process in terms of the original features. Furthermore, clustering the SHAP values has been proposed as a method for subgroup discovery \citep{ex1, ex2, ex3}. This supervised SHAP-based clustering method groups samples that not only received similar predictions, but also received similar predictions for similar reasons. A mapping of the heterogeneity in prediction can be leveraged to create data-driven solutions personalized to each subgroup.

Now the question arises -- How do the SHAP values differ for different machine learning models and how do these differences affect the SHAP analysis process? To answer these questions, we study three different machine learning models with varying levels of complexity across three different data sets. We also introduce the high-dimensional waterfall plot, a generalization of the classical waterfall plot to multi-classification problems, to help interpret cluster analyses of SHAP values.

\section{Background}

In this section, we describe existing methods used in our experiments.

\subsection{SHAP Analysis}

SHAP analysis \citep{shap} provides a representation of the learned relationships between predictors and response in terms of the original features. Given a sample, each feature is assigned an associated SHAP value quantifying the contribution of that feature towards the prediction of the given sample. Thus, for each sample, its prediction is allocated among the SHAP values for each individual feature, indicating the magnitude and direction of contribution of each feature. Unlike covariate analysis, SHAP analysis is conducted on a sample-by-sample basis. For one feature, there are individual SHAP values associated with each sample. In other words, the effect of each covariate varies sample to sample.

To provide a mathematical formulation, suppose $f:X \to \mathbb{R}$ is a trained model, where $f$ is trained on a data set $X' \subset X \subset \mathbb{R}^p$. In the case of a binary target variable, the model predicts the log-odds of success. For each $x \in X$, there exists a sequence of SHAP values $\phi(f;x)_1, \hdots, \phi(f;x)_p$, one associated to each feature, such that
\begin{equation*}
\sum_{i=1}^p \phi(f;x)_i = f(x) - \mathbb{E}[f(X')].
\end{equation*}
In other words, the SHAP values for a given sample sum to the difference between the output of that sample and the average output of the entire training set. The deviation of the sample’s prediction from the average prediction is allocated among the individual features.

It is important to note each SHAP value $\phi(f;x)_i$ can be positive or negative. When predicting log-odds, negative SHAP values contribute towards a prediction of failure, while positive SHAP values contribute towards a prediction of success. The significance of each feature in deciding the prediction is then proportional to the magnitude of the corresponding SHAP value.

In the case of multi-classification, the SHAP values are $k$-dimensional vectors, where $k$ is the total number of classes. Each component of a SHAP vector represents the contribution of the corresponding feature to the prediction of the corresponding class. Hence, the SHAP vectors can be organized in a $(n \times p \times k)$-dimensional tensor, where $n$ is the number of samples, $p$ is the number of features, and $k$ is the number of classes. To analyze, the last two dimensions can be flattened to form a $(n \times pk)$-dimensional matrix.

\subsection{Waterfall Plots}

Waterfall plots are designed to display SHAP explanations for individual samples. Like a bar chart, bars are used to represent the SHAP value for each feature. However, the bars are stacked so that the base of each bar is located at the tail of the previous bar. This way the bars represent the cumulative sum of the SHAP values. When the stack is anchored at the average prediction, the series of bars ends at the prediction of the given sample because the SHAP values sum to the difference between the sample's prediction and the average prediction. The bars are ordered by magnitude.

\begin{figure}[H]
\centering
\includegraphics[width=4in]{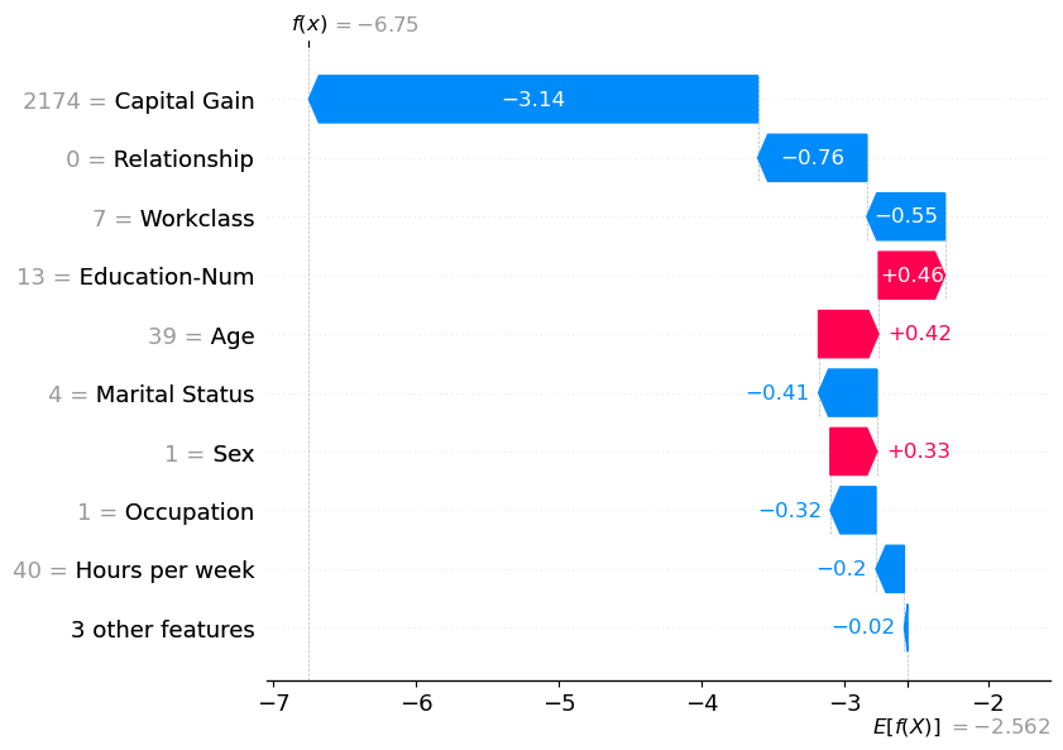}
\caption{Waterfall plot explaining the prediction of a specific individual.}
\label{waterfall_example}
\end{figure}

To provide an example, we explore the UCI adult income data set \citep{UCI_income}. The data contain demographic information and a predictive model is trained to predict whether an individual makes more or less than \$50K a year. Figure \ref{waterfall_example} depicts the waterfall plot for a specific individual.

Because the log-odds prediction for this individual is negative (-6.75), this individual is predicted to make less than \$50k a year. Moreover, the waterfall plot explains how this prediction is allocated among the features in the data. The amount of capital gains tax paid is by far the largest contributor towards the negative prediction for this individual. This individual did not pay any capital gains tax, meaning they did not make any profit selling capital assets, like stocks or real estate, in the previous year.

Waterfall plots, however, are only applicable to regression or binary classification problems, in which the SHAP values are 1-dimensional. In the multi-classification case, we have multi-dimensional SHAP vectors, as opposed to 1-dimensional SHAP values, which can cannot be represented by bars. We present a generalization of the waterfall plot to the multi-classification case in section 3.2.

\subsection{XGBoost}

XGBoost, or eXtreme Gradient Boosting \citep{xgboost}, is an algorithm based on gradient boosted decision trees developed to handle large and complex data sets efficiently. The concept behind gradient boosted decision trees is to sequentially train new decision trees that predict the residuals of the previous ensemble of trees. In doing so, each tree is correcting the errors of the previous ensemble of trees and polishing the complete model. The final ensemble of trees can then be used to accurately predict the output of new samples. Note, XGBoost can be applied to both regression and classification problems.

XGBoost is a popular choice when it comes to SHAP analysis. The tree ensemble architecture provides increased flexibility, allowing the model to learn more complex relationships than a simple decision tree. Moreover, TreeSHAP \citep{treeshap}, the algorithm for calculating SHAP values for tree-based models like XGBoost, is very efficient because of its ability to exploit tree-based architecture.

\subsection{Feedforward Neural Network}

A feedforward neural network (FFN) is a type of neural network in which the information flows in one direction. The data enter an input layer, pass through a series of hidden layers, then ultimately reach an output layer. Each layer consists of a set of neurons through which the data pass through. Notably, the number of neurons in the input layer must be equal to the number of features in the data, while the number of neurons in the output layer must be equal to the number of outputs. For example, if training a classifier to classify data into $k$ classes, then the output layer must contain $k$ neurons. Each neuron has an associated weight that must be calibrated during training, which is done by minimizing a loss function via gradient descent. The gradient of the loss function is calculated via a process called backpropagation, in which the error is propagated back through the network to calculate the partial derivatives with respect to each weight coefficient. Non-linearity is introduced into the system via activation functions. When information is passed between neurons, it first passes through a non-linear activation function. This non-linearity in the prediction process allows neural networks to learn and model complex data patterns.

Basic FFNs are very effective when the input data is independent. For more complex data, alterations of the architecture can improve performance. Convolutional neural networks (CNN) are a type of FNN that are preferred when classifying images. CNNs incorporate convolutional and pooling layers, which effectively reduce the number of features, allowing the network to detect multi-pixel structures such as curves and edges, as opposed to individual pixels. An ordinary FNN would required an impractically large number of neurons to digest an entire image pixel by pixel.

The calculation of SHAP values for neural networks is doable, but rather tedious. There is no dedicated algorithm for neural networks, so a sampling-based model-agnostic method must be used. Kernel SHAP \citep{shap} is the most popular choice.

\subsection{UMAP}

UMAP \citep{umap} belongs to a family of dimension reduction methods whose goal is to represent a high-dimensional data set using a fewer number of features. Being able to represent a data set using only two features allows us to plot the data on a two-dimensional set of axes for visualization. UMAP is the state-of-the-art dimension reduction method for visualization and cluster detection because of its prioritization of local structure. Its nonlinear nature also allows it to handle large reductions in the number of dimensions.

Notably, UMAP is known to produce false structures and over-cluster data when mishandled \citep{understanding_UMAP}. Care must be used when interpreting UMAP visualizations.

\subsection{HDBSCAN}

HDBSCAN \citep{hdbscan} is a hierarchical clustering method capable of handling noise and clusters with variable densities. Its predecessor, DBSCAN, is a non-hierarchical clustering method that struggles with clusters of variable densities. HDBSCAN solves this issue by implementing a hierarchical procedure at the cost of computational efficiency.

\section{Methods}

\subsection{Data}

\subsubsection{Simulated Data}

To test SHAP analysis's ability to detect and identify subgroups, we simulate data containing distinct pathways to the same target class. We assume the data follow a multinomial logistic regression model, i.e. the probabilities of belonging to each of three classes are linked to independent linear combinations of the input variables via the logistic function.

The coefficients are defined as follows,
\[f_1(x_1, \hdots, x_{10}) = 4x_1x_2 + 4x_1 + 4x_2 + \sum_{i=3}^{10} \beta_{1,i}x_i\]
\[f_2(x_1, \hdots, x_{10}) = 4x_1x_2 - 4x_1 - 4x_2 + \sum_{i=3}^{10} \beta_{2,i}x_i\]
where $\beta_{j,i} \overset{iid}{\sim} N(0, 1)$. With these coefficients, we expect Class 1 to contain the samples whose first two entries are positive and Class 2 to contain the samples whose first two entries are negative. The remainder of the samples, those whose first two entries are opposite signs, should belong to Class 3. There are two distinct pathways to arrive at Class 3: $x_1 > 0, x_2 < 0$ and $x_1 < 0, x_2 > 0$.

The input data are uniformly sampled from a hyperrectangle,
\begin{equation*}
X_1, \hdots, X_{1500} \overset{\textrm{iid}}{\sim} \textrm{Unif}([-5, 5]^{10}).
\end{equation*}
We sample 1,500 points in 10 dimensions. Then for each sample point $X_i$, we calculate the probabilities of belonging to each class
\[p_{1,i} = \frac{\exp(f_1(X_i))}{1 + \exp(f_1(X_i)) + \exp(f_2(X_i))}\]
\[p_{2,i} = \frac{\exp(f_2(X_i))}{1 + \exp(f_1(X_i)) + \exp(f_2(X_i))}\]
\[p_{3,i} = \frac{1}{1 + \exp(f_1(X_i)) + \exp(f_2(X_i))}\]
Lastly, the final class labels are sampled from a discrete distribution with the associated probabilities to introduce noise,
\[y_i = \begin{cases}
1 \textrm{ with probability } p_{1,i} \\
2 \textrm{ with probability } p_{2,i} \\
3 \textrm{ with probability } p_{3,i}
\end{cases}.\]

\subsubsection{MNIST Data}

The MNIST database \citep{MNIST} consists of $(28 \times 28)$-pixel images of handwritten digits with pixel intensities ranging from 0 to 255. Each image was flattened into a vector of length $28 \times 28 = 784$ for analysis.

\subsubsection{ADNI Data}

The Alzheimer's Disease Neuroimaging Initiative (ADNI) maintains a collection of longitudinal clinical, imaging, genetic, and other biomarker data. The ADNI was launched in 2003 as a public-private partnership, led by Principal Investigator Michael W. Weiner, MD. The primary goal of ADNI has been to test whether serial magnetic resonance imaging (MRI), positron emission tomography (PET), other biological markers, and clinical and neuropsychological assessment can be combined to measure the progression of mild cognitive impairment and early Alzheimer’s disease. For up-to-date information, see www.adni-info.org.

The various Alzheimer's data sets were combined using the ADNIMERGE R package \citep{adnimerge}. Visit data (visit code, exam date, site, study protocol), categorical demographic information (sex, ethnicity, marriage status), and equipment details (MRI Field Strength, FreeSurfer Software Edition, LONI image ID) were removed. The features missing readings for more than half of the patients were also removed (seven in total).

The cleaned data set contains 2,422 patients and 39 features. Each column was linearly re-scaled to range from zero to one to account for varying units of measurement. The target variable is comprised of three classes -- Cognitively Normal (CN), Mild Cognitive Impairment (MCI), and Alzheimer's disease (AD).

\subsection{High-Dimensional Waterfall Plots}

As discussed in section 2.2, the classical waterfall plot is only applicable to regression and binary classification problems. In the multi-classification case, the SHAP vectors cannot be represented by bars. To generalize the waterfall plot to multi-classification, we can, instead, represent the explanation of a sample with a high-dimensional path extruding from the origin. The path is constructed by concatenating the SHAP vectors corresponding to each feature in order of magnitude. Hence, for each sample, we construct a $k$-dimensional path, where $k$ is the number of classes, representing the allocation of its prediction among the individual features.

\begin{figure}[H]
\centering
\includegraphics[width=5in]{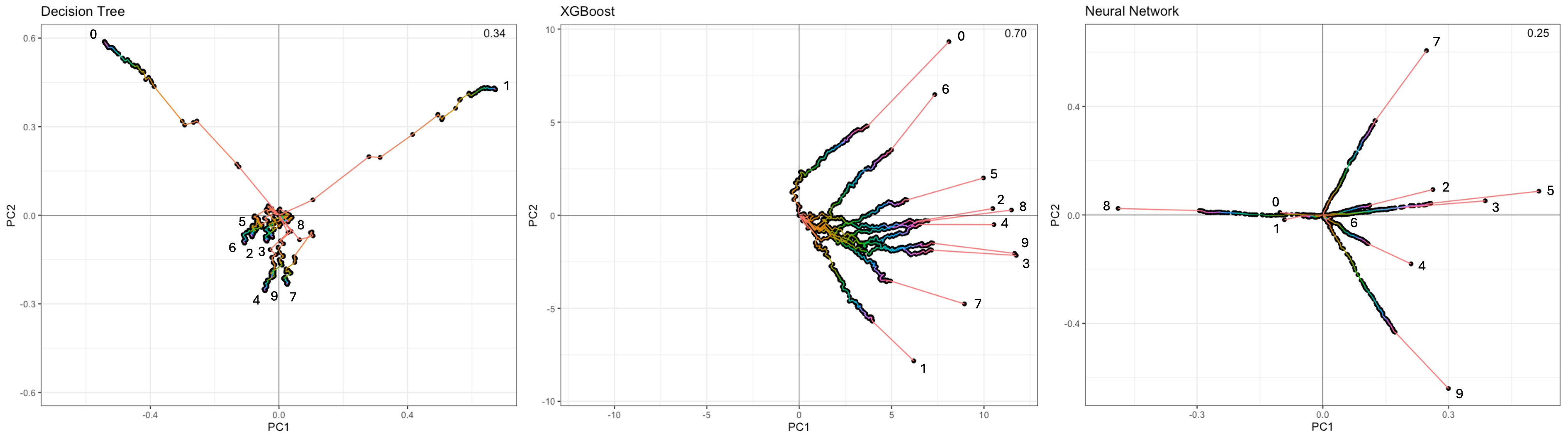}
\caption{High-dimensional clustered waterfall plots of SHAP vectors averaged across digit (MNIST).}
\label{waterfall_mnist}
\end{figure}

To visualize these paths, we project the paths onto the two-dimensional subspace that retains the most information using Principal Component Analysis (PCA). By visualizing the paths, we are able to understand the prediction of each sample in terms of the original features. The direction and magnitude of each segment explain the effect of the corresponding feature, while the total prediction is represented by the endpoint of each path. See Figures \ref{waterfall_sim}, \ref{waterfall_mnist}, and \ref{waterfall_adni} for examples.

It is also worth noting that the waterfall analysis, both the classical version and the high-dimensional version, can be conducted on a cluster-by-cluster basis, rather than a sample-by-sample basis, by averaging the SHAP values or vectors across clusters. We refer to the corresponding plot as the clustered waterfall plot.

\section{Results}
 
\subsection{Simulated Data}
 
Recall the simulated data was designed to contain samples assigned to the same class through distinct pathways. Because these pathways are built into the class-assignment function and not the raw data, these pathways are invisible prior to the SHAP analysis.

\begin{figure}[H]
\centering
\includegraphics[width=3in]{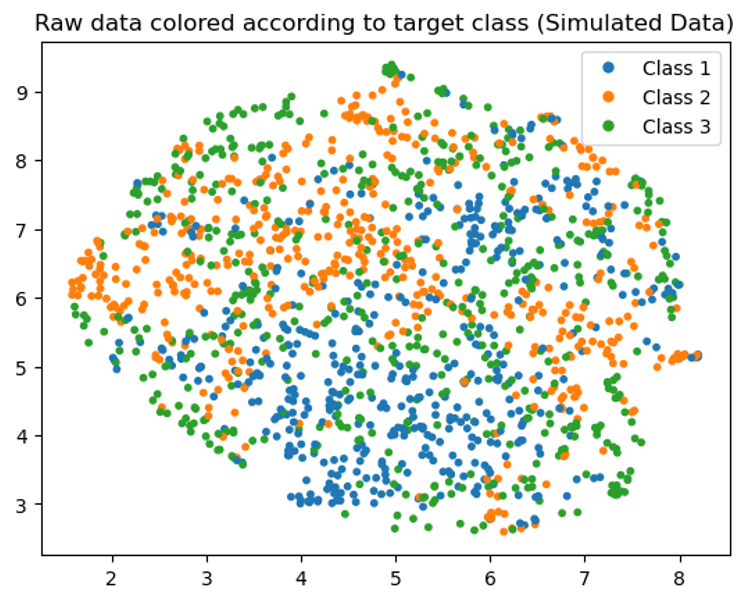}
\caption{Raw simulated data embedded in two dimensions with UMAP colored by target class.}
\label{raw_sim}
\end{figure}

When visualizing the raw simulated data, there are no meaningful clusters (Figure \ref{raw_sim}). The classes are indistinguishable by construction.
 
To better understand how SHAP analysis constructs model explanations, three different classifiers were trained -- a decision tree, an XGBoost model, and a feedforward neural network. Each model was trained on a randomly sampled training set containing 70\% of the data. Model performance was measured on a test set containing the remaining 30\% of the data. Hyperparameter optimization was conducting using grid search cross validation. Table \ref{sim_performance} contains the performance metrics for all three classifiers. All three classifiers exhibited adequate performance with XGBoost and the neural network outperforming the decision tree.

\begin{table}[H]
\centering
\begin{tabular}{ccccccccccc}
\toprule
& \multicolumn{2}{c}{Decision Tree} & \multicolumn{2}{c}{XGBoost} & \multicolumn{2}{c}{Neural Network} & \\
\cmidrule{2-7}
Class & Precision & Recall & Precision & Recall & Precision & Recall & Support\\
\midrule
1 & 0.87 & 0.93 & 0.95 & 0.96 & 0.94 & 0.97 & 138\\ 
2 & 0.86 & 0.89 & 0.95 & 0.97 & 0.96 & 0.96 & 142\\
3 & 0.94 & 0.86 & 0.96 & 0.93 & 0.96 & 0.94 & 170\\
\midrule
Accuracy & & 0.89 & & 0.94 & & 0.96 & 450\\
\bottomrule
\end{tabular}
\caption{\label{sim_performance}Classifier performance on simulated data}
\end{table}

\begin{figure}[H]
\centering
\includegraphics[width=5in]{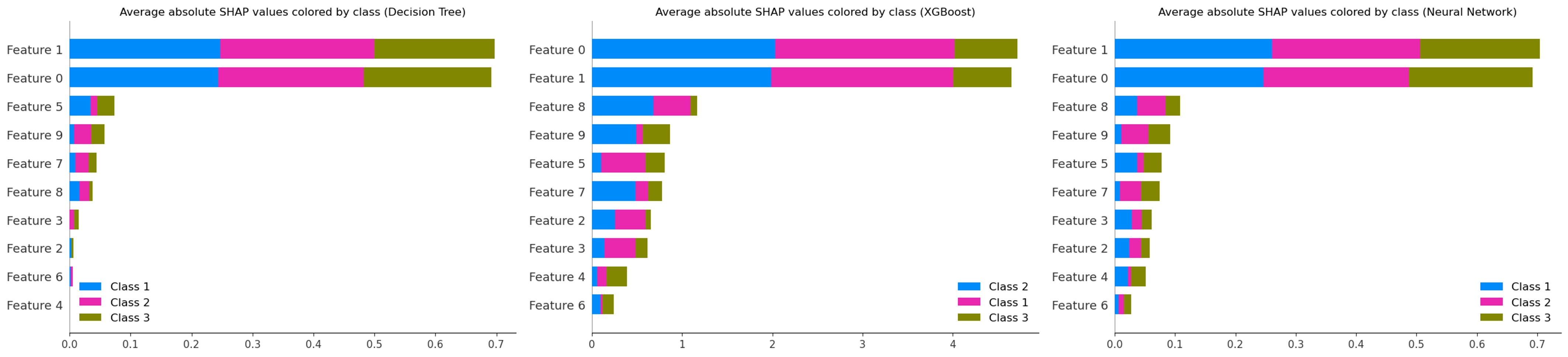}
\caption{Stacked bar chart of average absolute SHAP values for the simulated data.}
\label{bar_sim}
\end{figure}

SHAP analysis was conducted on all three classifiers. For each classifier, a stacked bar chart of average absolute SHAP values indicates which features were most influential (Figure \ref{bar_sim}). Unsurprisingly, Features 0 and 1 were the largest contributors to prediction in all three classifiers, as the remaining features were all noise by construction.

\begin{figure}[H]
\centering
\includegraphics[width=5in]{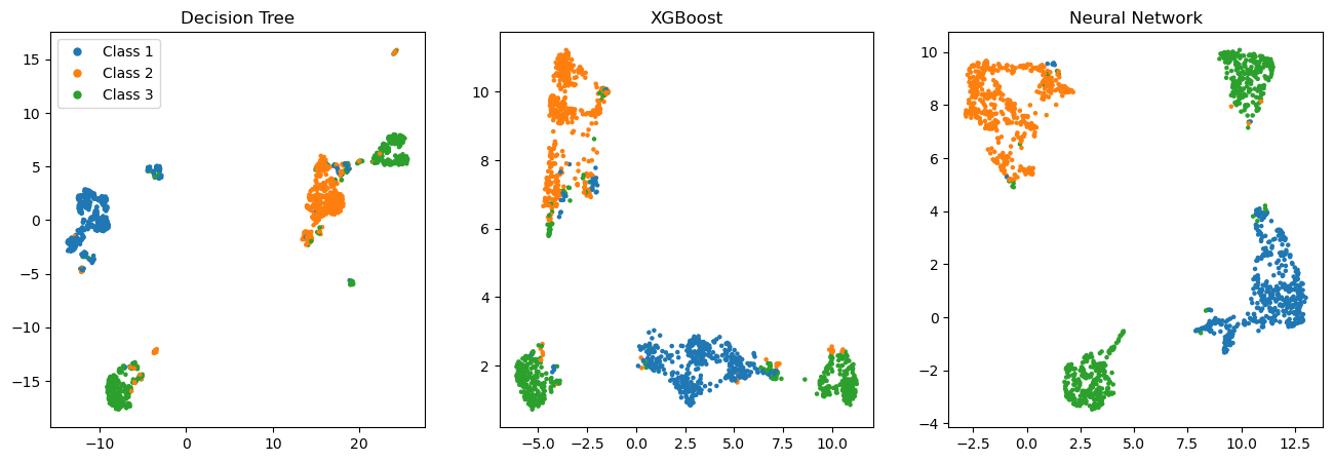}
\caption{UMAP embedding of SHAP vectors for all three classifiers trained on the simulated data.}
\label{shap_sim}
\end{figure}

Perhaps even more indicative of the difference in the three classifiers is a visualization of the SHAP vectors. UMAP embeddings of the flattened SHAP vectors for each classifier provide more insight into the model explanations (Figure \ref{shap_sim}).

The SHAP vectors for the decision tree cluster into four major clusters -- one cluster each for Classes 1 and 2 and two clusters corresponding to Class 3. Scattered around the major clusters are smaller clusters containing a few points each. The sporadic nature of the clusters is representative of the decision tree's discrete structure. Each prediction is a series of binary decisions terminating at one of finitely many leaf nodes. If we color the SHAP vectors by leaf node, we see each cluster corresponds to one of the many leaf nodes (Figure S1).

The SHAP vectors for XGBoost and the neural network also depict the same four major clusters. Unlike the SHAP vectors for the decision tree, these SHAP vectors do not also cluster into numerous minor clusters. Furthermore, the neural network SHAP vectors seem to have slightly fewer misclassified points within each class, reflecting the neural network's marginally superior performance.

Clearly, Class 3 contains two distinct subgroups. To interpret these subgroups, we conduct a subgroup analysis.

\subsubsection{Subgroup Discovery}

\begin{figure}[H]
\centering
\includegraphics[width=5in]{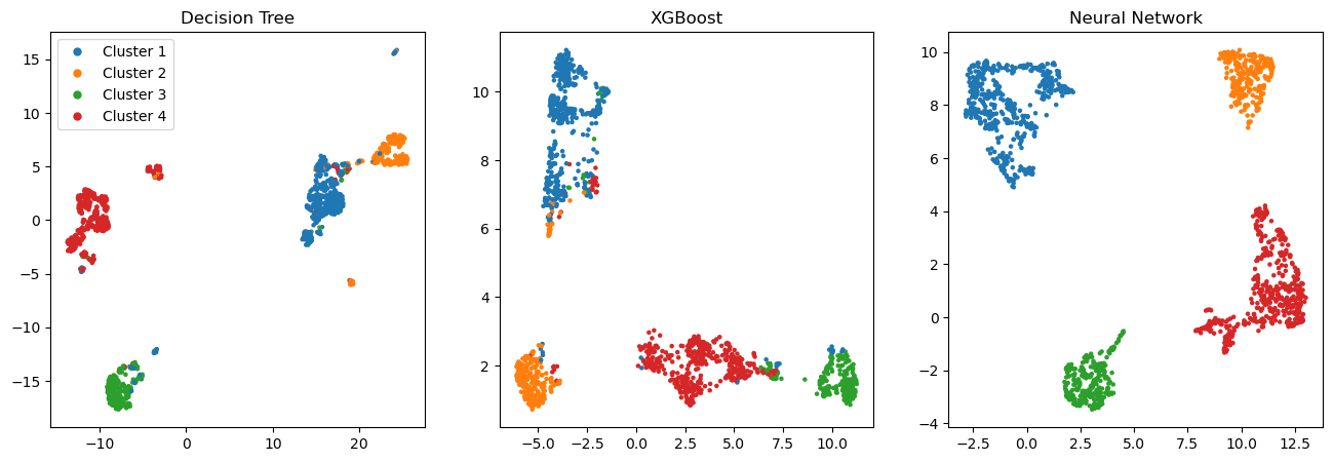}
\caption{SHAP vectors colored according to an HDBSCAN cluster analysis of the XGBoost SHAP vectors (simulated data).}
\label{hdbscan_sim}
\end{figure}

The SHAP vectors for all three classifiers capture two distinct subgroups of Class 3. We study the neural network SHAP vectors because they exhibit the cleanest clustering. A cluster analysis was conducted on the neural network SHAP vectors then applied to the SHAP vectors for all three classifiers (Figure \ref{hdbscan_sim}). The cluster analysis was conducted using HDBSCAN, in line with past articles also conducting SHAP-based clustering \citep{ex1, ex3}. All three classifiers seem to agree on the subgrouping.

\begin{figure}[H]
\centering
\includegraphics[width=5in]{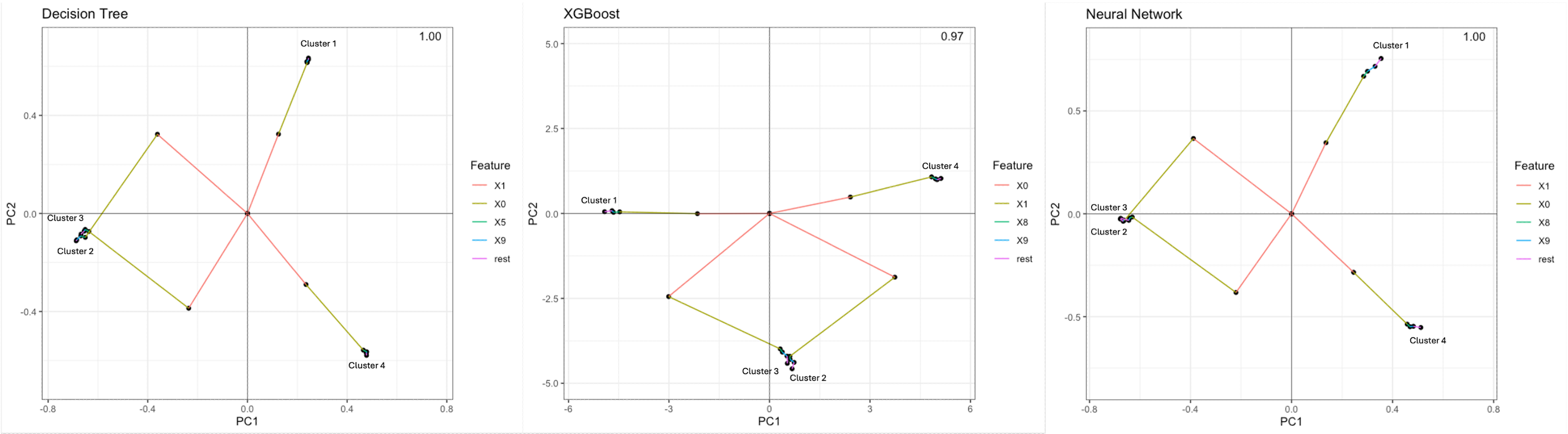}
\caption{High-dimensional clustered waterfall plots of SHAP vectors averaged across HDBSCAN clustering (simulated data).}
\label{waterfall_sim}
\end{figure}

To better understand the inter-cluster relationships, we use the high-dimensional clusterered waterfall plot. By averaging the SHAP vectors across each cluster then ordering the average vectors by magnitude, we create feature-by-feature path representations of each cluster. In particular, each cluster is represented by a series of average SHAP vectors that are concatenated to form a high-dimensional path. Each path gives a feature-by-feature explanation of how the average sample reached its prediction, which is represented by the endpoint of the path. PCA projections of these paths are given in Figure \ref{waterfall_sim}. The plots for each classifier tell a similar story. Features 0 and 1 dominate the predictions in all three classifiers. Moreover, Clusters 2 and 3 received similar predictions for distinct reasons, as illustrated by their distinct pathways to nearby endpoints.

\begin{figure}[H]
\centering
\includegraphics[width=3in]{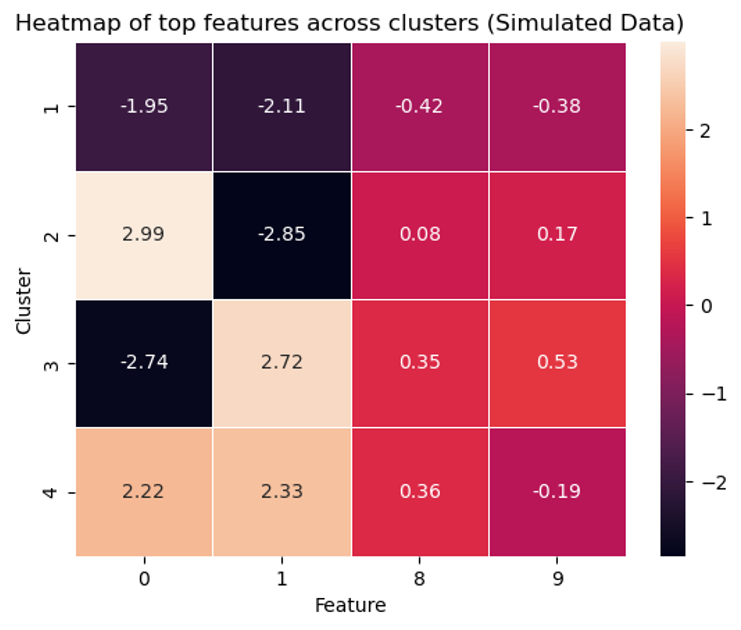}
\caption{Heat map of raw data averaged across clusters (simulated data).}
\label{heatmap_sim}
\end{figure}

We can further interpret the clusters using a heat map of the raw data. Figure \ref{heatmap_sim} illustrates the values of the most influential features averaged across clusters. Notably, Cluster 1 contains the samples whose first two features are negative, Cluster 4 contains the samples whose first two features are positive, and Clusters 2/3 contain the samples whose first two features are opposite signs. The two distinct pathways portrayed in the high-dimensional clustereted waterfall plot are a positive Feature 0 followed by a negative Feature 1 (Cluster 2) and a negative Feature 0 followed by a positive Feature 1 (Cluster 3).

Now, if we refer back to the high-dimensional clustered waterfall plots, we see these inter-cluster relationships are captured by the plotted paths. Consider Cluster 3 in the plot for the decision tree, for example. Its Feature 0 segment runs in the opposite direction of the Cluster 1 path, and its Feature 1 segment runs in the opposite direction of the Cluster 4 path. In particular, Cluster 3 is differentiated from Cluster 1 via Feature 0 and differentiated from Cluster 4 via Feature 1. Both the SHAP-based clustering and the high-dimensional clustered waterfall plot were able to model the subgroups defined by the class-assignment function.

\subsection{MNIST}

Image classification is a very complex task. Different classifiers process images in very different ways. Through this example, we illustrate SHAP analysis's ability to capture the distinct decision-making processes of different classifiers.

\begin{figure}[H]
\centering
\includegraphics[width=3in]{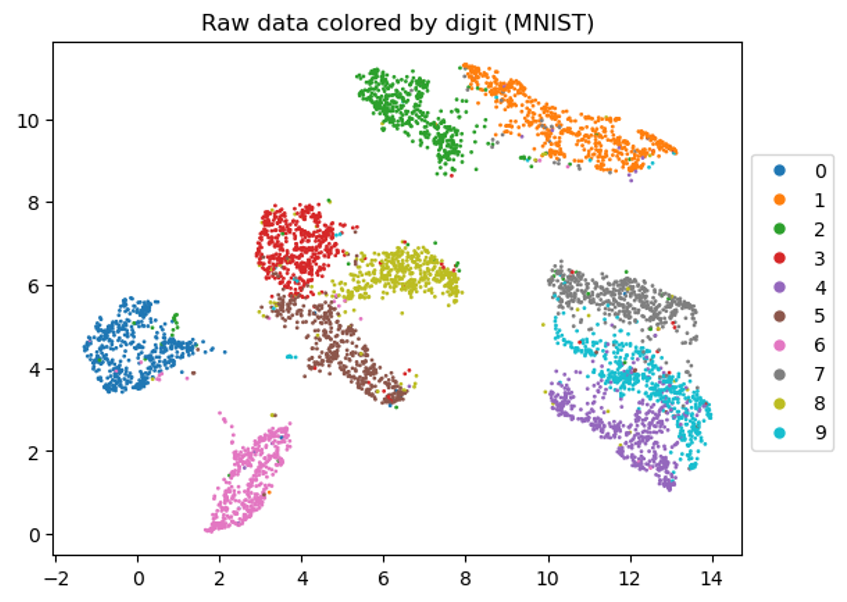}
\caption{Raw MNIST data embedded in two dimensions with UMAP colored by digit.}
\label{raw_mnist}
\end{figure}

25,000 images were randomly sampled from the MNIST data set. Of them, 20,000 were allocated to a training set and the remaining 5,000 were allocated to a test set. Visualizing the raw test set, we already see some separation between digits (Figure \ref{raw_mnist}). UMAP, however, has a tough time distinguishing between digits 3, 8, and 5 as well as 4, 7, and 9. A predictive model is needed to detect the fine distinction between these sets of digits.

A decision tree, an XGBoost model, and a CNN were trained on the training data. Their test set performance metrics are given in Table \ref{mnist_performance}. Unsurprisingly, the decision tree performed the worst, while the CNN performed the best. Digits are defined by complex inter-pixel structures like curves and edges, which are best learned by a CNN.

\begin{table}[H]
\centering
\begin{tabular}{ccccccccccc}
\toprule
 & \multicolumn{2}{c}{Decision Tree} & \multicolumn{2}{c}{XGBoost} & \multicolumn{2}{c}{Neural Network} & \\
 \cmidrule{2-7}
Digit & Precision & Recall & Precision & Recall & Precision & Recall & Support\\
\midrule
0 & 0.89 & 0.92 & 0.96 & 0.98 & 0.99 & 0.98 & 503\\
1 & 0.94 & 0.92 & 0.98 & 0.99 & 1.00 & 1.00 & 577\\ 
2 & 0.77 & 0.78 & 0.97 & 0.94 & 0.98 & 0.98 & 524\\
3 & 0.74 & 0.68 & 0.95 & 0.95 & 1.00 & 0.98 & 513\\
4 & 0.68 & 0.80 & 0.97 & 0.97 & 0.99 & 0.98 & 481\\
5 & 0.69 & 0.71 & 0.95 & 0.95 & 0.98 & 0.99 & 434\\
6 & 0.85 & 0.84 & 0.95 & 0.97 & 0.98 & 0.99 & 457\\
7 & 0.86 & 0.84 & 0.97 & 0.95 & 0.98 & 0.97 & 511\\
8 & 0.80 & 0.74 & 0.95 & 0.93 & 0.99 & 0.98 & 490\\
9 & 0.75 & 0.73 & 0.93 & 0.95 & 0.97 & 0.99 & 510\\
\midrule
Accuracy & & 0.80 & & 0.96 & & 0.98 & 5000\\
\bottomrule
\end{tabular}
\caption{\label{mnist_performance}Classifier performance on MNIST data.}
\end{table}

\begin{figure}[H]
\centering
\includegraphics[width=5in]{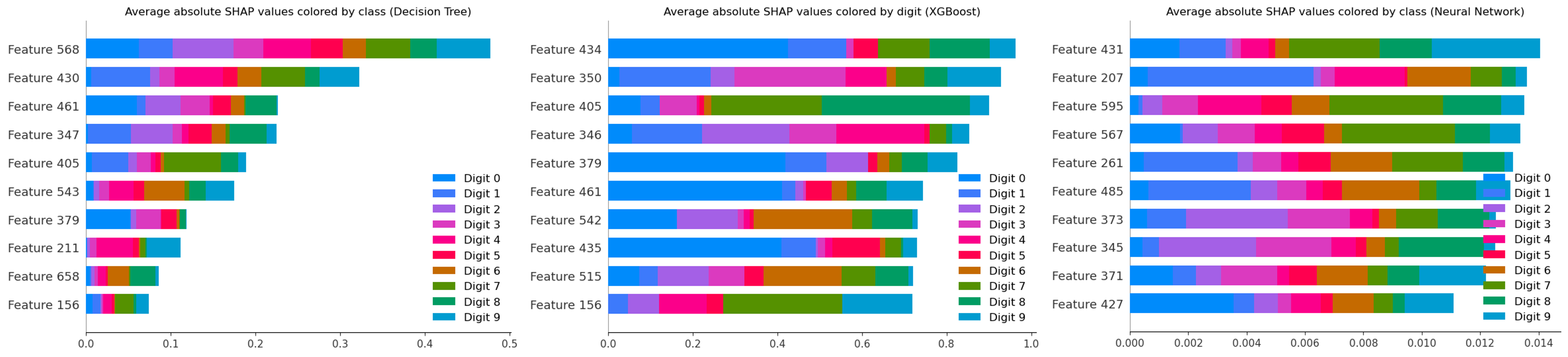}
\caption{Stacked bar chart of average absolute SHAP values for the MNIST data.}
\label{bar_mnist}
\end{figure}

Figure \ref{bar_mnist} shows the stacked bar charts of average absolute SHAP values. Among the three classifiers, there was no agreement on which features were most useful because individual pixels are not enough to distinguish digits. This phenomenon is corroborated by the near-equal distribution of SHAP values among the features for XGBoost and the CNN.

\begin{figure}[H]
\centering
\includegraphics[width=5in]{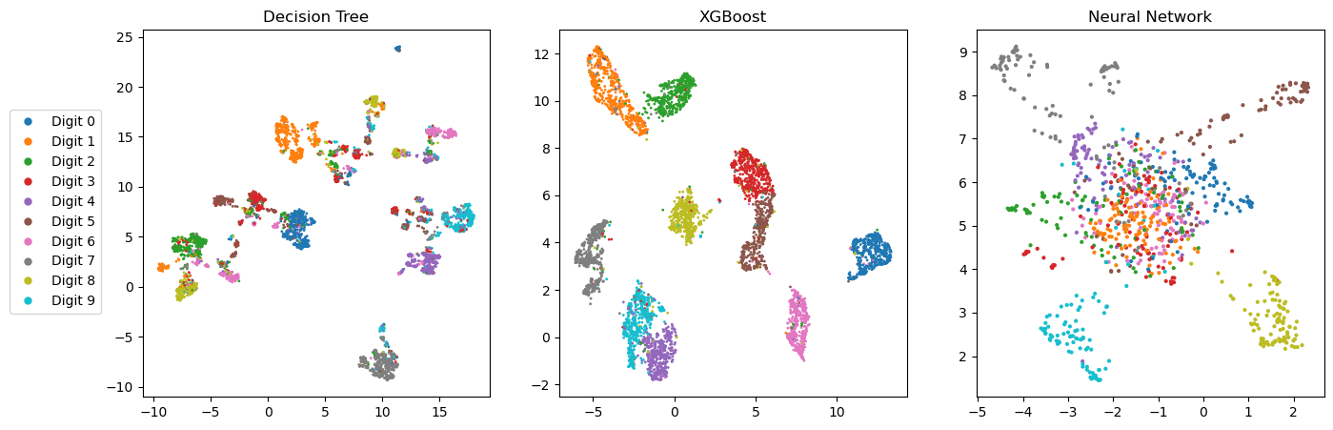}
\caption{UMAP embedding of SHAP vectors for all three classifiers trained on the MNIST data.}
\label{shap_mnist}
\end{figure}

The UMAP embeddings of SHAP vectors (Figure \ref{shap_mnist}) provide more insight on the decision-making processes behind the three classifiers. The SHAP vectors for the decision tree are, again, very sporadic and clustered according to leaf node (Figure S2). The SHAP vectors for XGBoost cluster according to digit, but unlike the raw values, digit 8 is separated from digits 3 and 5, and digit 7 is separated from digits 4 and 9. The SHAP vectors for the CNN form clusters corresponding to digits 5, 7, 8, and 9, while the remaining points form another amalgamated cluster.

The SHAP vectors for the CNN are particularly interesting. They suggest some of the digits are rather difficult to distinguish, and yet, the CNN exhibited the highest prediction accuracy. This inconsistency is due to the incompatibility of SHAP's additive nature and the CNN's decision-making process. SHAP analysis assumes prediction can be modeled as a sum of SHAP values, one corresponding to each feature, or pixel in this case. For image analysis, CNNs are designed to capture multi-pixel, location-invariant structures like curves and edges. Because these structures are not defined by their location in the image, the SHAP vectors are not consistent among images of the same digit. The same stroke in different images of the same digit are not defined by the same set of pixels, so the misaligned SHAP vectors do not cluster nicely.

The high-dimensional clustered waterfall plot provides additional context on the SHAP vectors , especially for the CNN. Like the UMAP embedding of CNN SHAP vectors, the waterfall plot depicts digits 5, 7, 8, and 9 as the most defined digits, but we can better see how the other digits compare:
\begin{itemize}
	\item 0 and 1 are similar to 8,
	\item 2, 3, and 6 are similar to 5, and
	\item 4 is similar to 9.
\end{itemize}
All of these relationships are consistent with how the digits are written.

\subsection{ADNI}

As model complexity increases, model interpretability decreases. This decreases the applicability of complex machine learning models in trust-sensitive fields like medicine. SHAP analysis and other explainable AI methods alleviate this problem by explaining and contextualizing the decision-making processes of these black-box models. Through this example, we demonstrate how interpretable, data-driven solutions can be extracted from the SHAP vectors.

The cleaned ADNI data set contains 2,422 patients. However, UMAP and the trained neural network are unable to account for missing data. Among the 2,422 patients, 737 of them contain no missing data. For these two sets of patients, the data was allocated to a training set and test set according to a 70\%/30\% split.

\begin{figure}[H]
\centering
\includegraphics[width=3in]{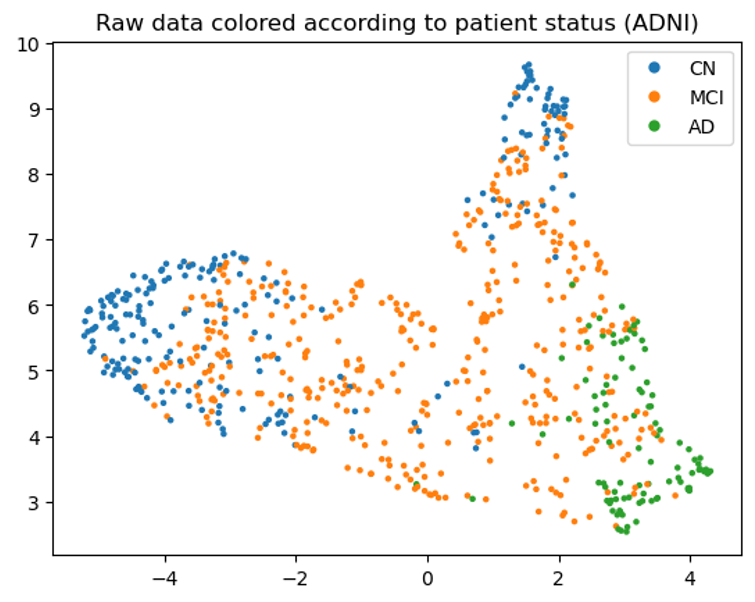}
\caption{Raw ADNI data embedded in two dimensions with UMAP colored by patient status.}
\label{raw_adni}
\end{figure}

The raw ADNI data cluster according to patient status, but the decision boundaries are muddled (Figure \ref{raw_adni}). Moreover, there are two distinct subgroups of cognitively normal patients.

A decision tree, an XGBoost model, and a neural network were trained on the training data. Their test set performance metrics are given in Table \ref{adni_performance}. All three classifiers exhibited adequate performance. 

\begin{table}[H]
\centering
\begin{tabular}{ccccccccc}
\toprule
& \multicolumn{2}{c}{Decision Tree} & \multicolumn{2}{c}{XGBoost} & & \multicolumn{2}{c}{Neural Network} & \\
\cmidrule{2-5}\cmidrule{7-8}
Class & Precision & Recall & Precision & Recall & Support & Precision & Recall & Support\\
\midrule
CN & 0.99 & 0.96 & 0.99 & 0.95 & 257 & 0.94 & 0.93 & 71\\
MCI & 0.88 & 0.96 & 0.91 & 0.96 & 342 & 0.93 & 0.90 & 118\\ 
AD & 0.90 & 0.73 & 0.90 & 0.83 & 128 & 0.79 & 0.91 & 33\\
\midrule
Accuracy & & 0.92 & & 0.93 & 727 & & 0.91 & 222\\
\bottomrule
\end{tabular}
\caption{\label{adni_performance}Classifier performance on ADNI data.}
\end{table}

\begin{figure}[H]
\centering
\includegraphics[width=5in]{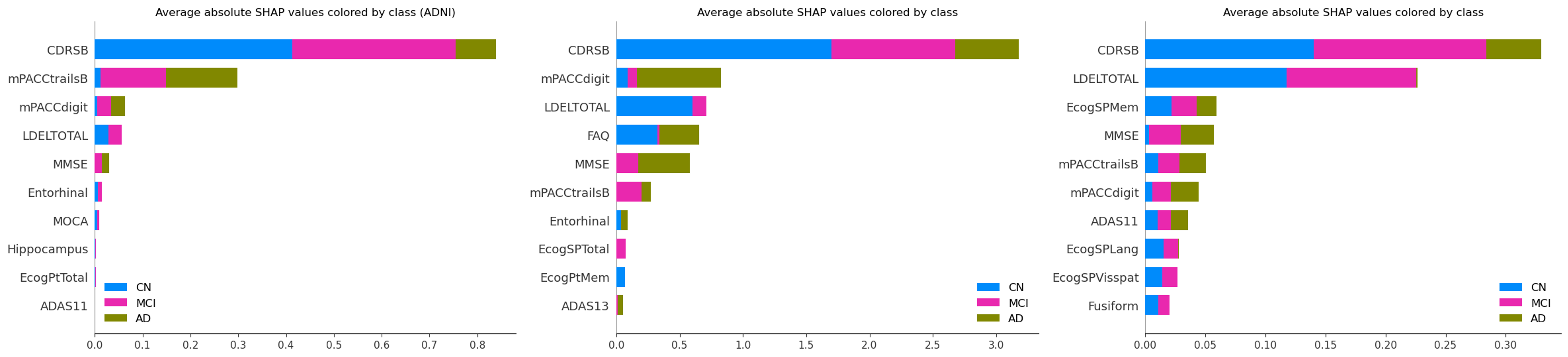}
\caption{Stacked bar chart of average absolute SHAP values for the ADNI data.}
\label{bar_adni}
\end{figure}

The stacked bar charts of average absolute SHAP values (Figure \ref{bar_adni}) all list CDRSB\footnote{Clinical Dementia Rating \citep{CDSRB} is a measure of cognition obtained by interviewing both the patient and a care partner. The questionnaire measures six different domains including memory and problem-solving. CDRSB is the sum of boxes, or sum of scores, across the six domains. Higher scores indicate cognitive impairment.} as the most influential feature in prediction. The subsequent most influential features differ between classifiers.

\begin{figure}[H]
\centering
\includegraphics[width=5in]{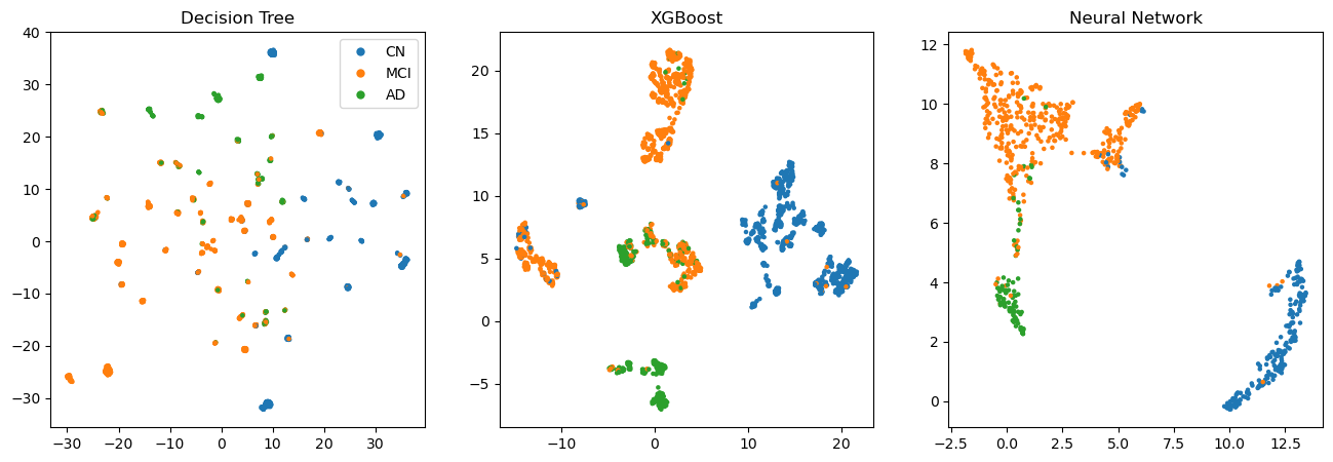}
\caption{UMAP embedding of SHAP vectors for all three classifiers trained on the ADNI data.}
\label{shap_adni}
\end{figure}

The UMAP embeddings of the SHAP vectors tell three different stories (Figure \ref{shap_adni}). Like the previous two examples, the SHAP vectors for the decision tree cluster according to leaf node (Figure S3). The SHAP vectors for XGBoost are separated by patient status, but there exist numerous subgroups within each class. The SHAP vectors for the neural network do no exhibit this subgrouping. Instead, they depict a continuous progression from MCI to AD and a detached CN cluster.

To determine if these subgroups hold any meaning, we conduct a subgroup analysis.

\subsubsection{Subgroup Discovery}

\begin{figure}[H]
\centering
\includegraphics[width=5in]{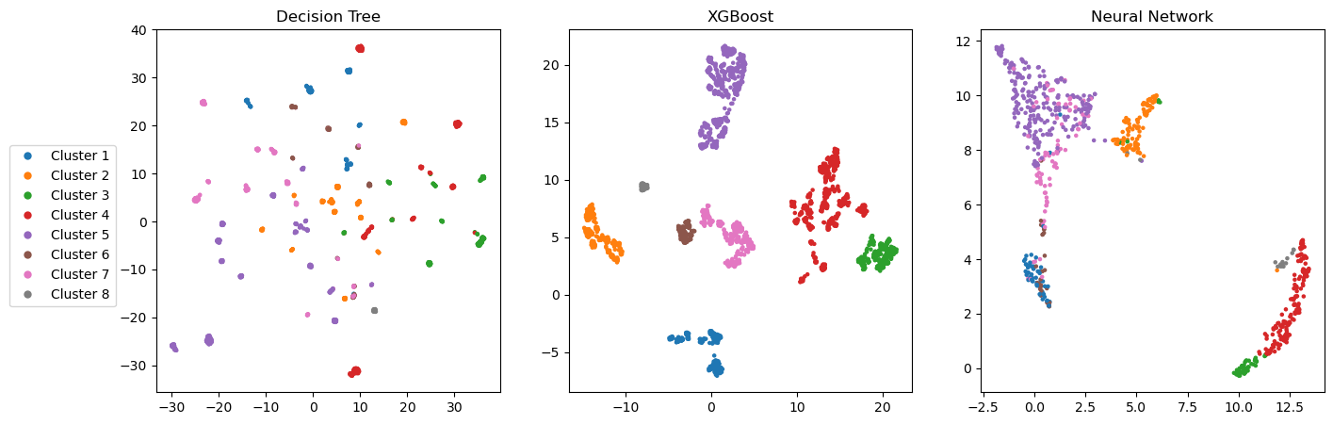}
\caption{SHAP vectors colored according to an HDBSCAN cluster analysis of the XGBoost SHAP vectors (ADNI).}
\label{hdbscan_adni}
\end{figure}

The SHAP vectors for XGBoost cluster into the finest clustering, so we investigate those subgroups. An HDBSCAN clustering was calculated upon the XGBoost SHAP vectors then applied to the SHAP vectors of all three classifiers (Figure \ref{hdbscan_adni}). The neural network SHAP vectors seem to preserve the subgrouping, but subgroups are no longer separated.

\begin{figure}[H]
\centering
\includegraphics[width=5in]{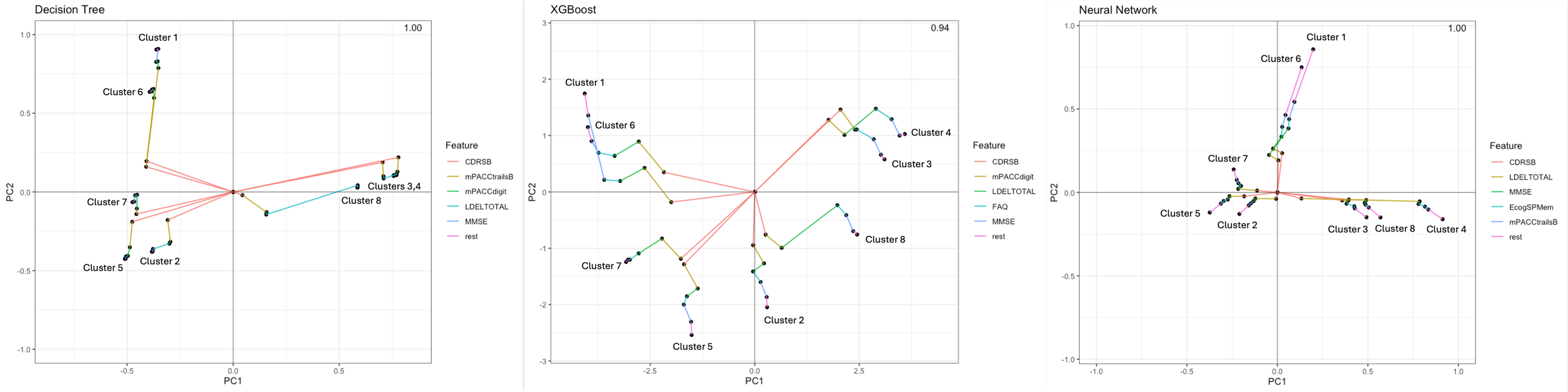}
\caption{High-dimensional clustered waterfall plots of SHAP vectors averaged across HDBSCAN clustering (ADNI).}
\label{waterfall_adni}
\end{figure}

The high-dimensional clustered waterfall plot provides a more in-depth view of the inter-cluster relationships (Figure \ref{waterfall_adni}). In all three plots, the paths partition into three separate groups: Clusters 3, 4, and 8; Clusters 2, 5, and 7; and Clusters 1 and 6, which correspond to CN, MCI, and AD patients, respectively. This is a more accurate representation of the global inter-cluster relationships because UMAP is known to struggle with global cluster positioning \citep{understanding_UMAP}. This is why the UMAP embedding of the neural network SHAP vectors better captures this grouping than the UMAP embedding of the XGBoost SHAP vectors. Once the groups are broken up into smaller clusters, their global positioning is distorted by UMAP.

\subsubsection{Data-Driven Solutions}

By analyzing these paths, we gain a better understanding of the heterogeneity in prediction, which can be leveraged to create data-driven solutions. For example, Clusters 1 and 6 contain the patients with Alzheimer's disease. However, the patients in each cluster were predicted to be cognitively impaired for distinct reasons, suggesting treatment personalized to each subgroup may be more effective than a catch-all remedy. If we examine the SHAP vectors for XGBoost, the best-performing model, we see the largest discrepancy between Clusters 1 and 6 are the features CDRSB and MMSE\footnote{The Mini-Mental State Examination is an 11-question measure that tests five areas of cognitive function: orientation, registration, attention and calculation, recall, and language. Lower scores indicate cognitive impairment.}. The SHAP values (Table \ref{adni_treatment}) suggest CDRSB is more influential in the predictions of patients belonging to Cluster 6, while MMSE is more influential in the predictions of patients belonging to Cluster 1.

\begin{table}[H]
\centering
\begin{tabular}{ccccccc}
\toprule
& \multicolumn{3}{c}{CDRSB} & \multicolumn{3}{c}{MMSE}\\
\cmidrule{2-4}\cmidrule{5-7}
Cluster & CN & MCI & AD & CN & MCI & AD\\
\midrule
1 & -2.03 & -0.19 & 0.58 & 0 & -0.87 & 0.84\\
6 & -2.03 & 0.59 & 0.97 & 0 & -0.02 & 0.31\\
\bottomrule
\end{tabular}
\caption{\label{adni_treatment}Average SHAP values for Clusters 1 and 6 calculated from the XGBoost model.}
\end{table}

Using this information, one might hypothesize treatments targeting the six domains measured by Clinical Dementia Rating will be more effective in Cluster 6, while treatments targeting the five areas of cognitive function measured by the Mini-Mental State Examination will be more effective in Cluster 1.

\section{Discussion}

Through these three examples, we showcased SHAP analysis's ability to capture the decision-making processes of various machine learning models. The increased complexity as you move from decision trees to XGBoost to neural networks has, both intended and unintended, effects on the SHAP values. A better understanding of these subtleties empowers analysts to get more from their SHAP analyses.

Decision trees are constructed via a finite series of binary decisions through which predictions are given by the leaf nodes. This discrete nature is reflected in the SHAP values. The SHAP values form sparse clusters corresponding to the numerous leaf nodes in the decision tree architecture.

XGBoost is a gradient-boosted decision tree model that aims to smooth the decision-making process by iteratively training new trees to correct the previous ensemble of trees. In this way, it acts as a discrete approximation of a continuous, or infinite, decision-making process. As such, the SHAP values calculated from an XGBoost model are not as sparse and discrete as those calculated from a single decision tree. While the tree structure still encourages clustering, the SHAP values form larger and more cohesive clusters than their decision tree counterparts, making XGBoost a great candidate for subgroup discovery. Some care, however, must be taken when visualizing these subgroups. Nonlinear dimension reduction methods like UMAP tend to over-cluster and struggle with cluster positioning. The clustered waterfall plot provides a much more reliable description of inter-subgroup relationships since it uses PCA, a linear dimension reduction method, to visualize cluster positioning. It is also worth noting calculating the SHAP values for XGBoost and other tree-based models is more efficient than other models.

Neural networks represent maximal flexibility. The Universal Approximation Theorems state neural networks can serve as a universal approximator for any continuous function \citep{UATs}. Therefore, neural networks can incorporate the maximal level of flexibility into their decision-making process. The non-linearity in the decision-making process can lead to SHAP values that do not cluster as distinctly as those from a tree-based model but may be a better representation if the class-assignment function is highly complex. In some cases, however, the complexity of neural networks can lead to unexpected results. In our image analysis, for example, the strokes captured by the CNN did not always occur in the same locations, so the SHAP vectors were misaligned. This misalignment muddied the visualization of SHAP values.

Beyond illustrating how SHAP analysis captures decision-making processes of machine learning models, we also demonstrated how SHAP-based clustering can be used for subgroup discovery. SHAP analysis allows us to cluster according to model explanation, which can be more useful than simply clustering the raw data. In doing so, we discover prediction archetypes representing the distinct reasons samples received the predictions they did. These explanations can be leveraged to construct data-driven solutions unique to each archetype that are more effective than any general solution.


\clearpage
\section*{Acknowledgments}

Data collection and sharing for this project was funded by the Alzheimer's Disease Neuroimaging Initiative (ADNI) (National Institutes of Health Grant U01 AG024904) and DOD ADNI (Department of Defense award number W81XWH-12-2-0012). ADNI is funded by the National Institute on Aging, the National Institute of Biomedical Imaging and Bioengineering, and through generous contributions from the following: AbbVie, Alzheimer’s Association; Alzheimer’s Drug Discovery Foundation; Araclon Biotech; BioClinica, Inc.; Biogen; Bristol-Myers Squibb Company; CereSpir, Inc.; Cogstate; Eisai Inc.; Elan Pharmaceuticals, Inc.; Eli Lilly and Company; EuroImmun; F. Hoffmann-La Roche Ltd and its affiliated company Genentech, Inc.; Fujirebio; GE Healthcare; IXICO Ltd.; Janssen Alzheimer Immunotherapy Research \& Development, LLC.; Johnson \& Johnson Pharmaceutical Research \& Development LLC.; Lumosity; Lundbeck; Merck \& Co., Inc.; Meso Scale Diagnostics, LLC.;  NeuroRx Research; Neurotrack Technologies; Novartis Pharmaceuticals Corporation; Pfizer Inc.; Piramal Imaging; Servier; Takeda Pharmaceutical Company; and Transition Therapeutics. The Canadian Institutes of Health Research is providing funds to support ADNI clinical sites in Canada. Private sector contributions are facilitated by the Foundation for the National Institutes of Health (www.fnih.org). The grantee organization is the Northern California Institute for Research and Education, and the study is coordinated by the Alzheimer’s Therapeutic Research Institute at the University of Southern California. ADNI data are disseminated by the Laboratory for Neuro Imaging at the University of Southern California.

\section*{Author Contributions}
J.L. performed the research and wrote the manuscript. J.F. reviewed and verified the research. All authors reviewed the manuscript.

\section*{Conflicts of Interest}
The authors have no conflicts of interest to declare.


\clearpage
\bibliographystyle{chicago}
\bibliography{ref}

\end{document}